\documentclass[letterpaper, 10 pt, conference]{ieeeconf}

\IEEEoverridecommandlockouts
\usepackage{array}
\usepackage[T1]{fontenc}
\usepackage{cite}
\usepackage{amsmath,amssymb,amsfonts}
\usepackage{algorithmic}
\usepackage{graphicx}
\usepackage{textcomp}
\usepackage{xcolor}
\usepackage{caption}
\usepackage{algorithm,algorithmic}
\usepackage{lipsum}
\usepackage{gensymb}
\usepackage{mleftright}
\usepackage{bm}
\usepackage{soul}
\usepackage{comment}
\usepackage{mathrsfs}
\usepackage{multirow}
\usepackage{hyperref}
\renewcommand{\arraystretch}{1.2}

\usepackage{amsthm}
\usepackage[algo2e,ruled,vlined,linesnumbered,resetcount,norelsize]{algorithm2e}

\makeatletter
\newcommand{\linebreakand}{%
  \end{@IEEEauthorhalign}
  \hfill\mbox{}\par
  \mbox{}\hfill\begin{@IEEEauthorhalign}
}
\makeatother

\newtheorem{theorem}{Theorem}

\newtheorem{remark}{Remark}

\newcommand{\tr}{\text{Tr}}

\def\BibTeX{{\rm B\kern-.05em{\sc i\kern-.025em b}\kern-.08em
    T\kern-.1667em\lower.7ex\hbox{E}\kern-.125emX}}
\begin{document}


\title{Human-Robot Co-Transportation using Disturbance-Aware MPC with Pose Optimization\\
}
\author{
Al Jaber Mahmud, Amir Hossain Raj, Duc M. Nguyen, Weizi Li, Xuesu Xiao, and Xuan Wang
\thanks{A. Mahmud, A. Raj, D. Nguyen, X. Xiao, and X. Wang are with George Mason University. W. Li is with the University of Tennessee, Knoxville. Supplementary material of this work can be found at \cite{supp}. This work was supported by NSF ECCS Award 2332210.
} }

\maketitle
\begin{abstract}
This paper proposes a new control algorithm for human-robot co-transportation using a robot manipulator equipped with a mobile base and a robotic arm. 
We integrate the regular Model Predictive Control (MPC) with a novel pose optimization mechanism to more efficiently mitigate disturbances (such as human behavioral uncertainties or robot actuation noise) during the task.
The core of our methodology involves a two-step iterative design: At each planning horizon, we determine the optimal pose of the robotic arm (joint angle configuration) from a candidate set, aiming to achieve the lowest estimated control cost. This selection is based on solving a disturbance-aware Discrete Algebraic Ricatti Equation (DARE), which also determines the optimal inputs for the robot’s whole body control (including both the mobile base and the robotic arm). To validate the effectiveness of the proposed approach, we provide theoretical derivation for the disturbance-aware DARE and perform simulated experiments and hardware demos using a Fetch robot under varying conditions,  including different trajectories and different levels of disturbances. The results reveal that our proposed approach outperforms baseline algorithms. 
\end{abstract}

\section{Introduction}

\begin{figure}[t]
    \centering
    \includegraphics[width=0.48\textwidth]{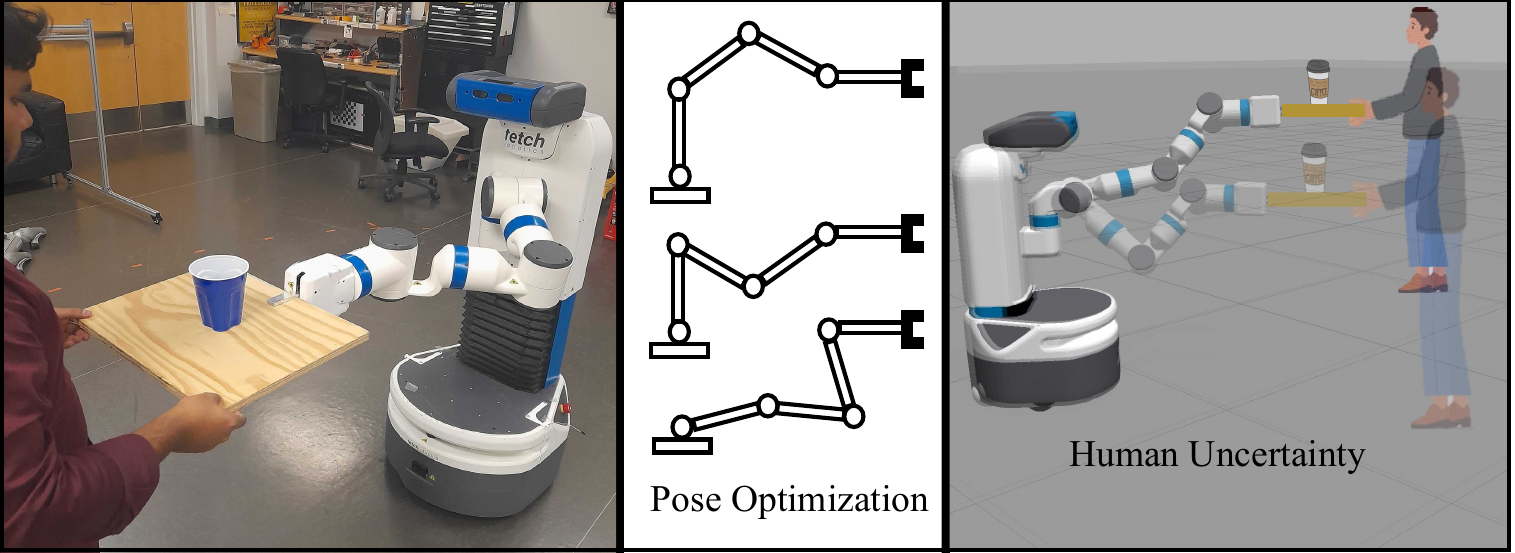}
    \caption{Human robot co-transportation through disturbance-aware MPC and pose optimization (redundancy resolution for the same end effector position).\vspace{-2em}}
    \label{fig_task}
\end{figure}

Collaborative human-robot systems can significantly reduce human workloads (Fig.~\ref{fig_task}). The capability of autonomous robots to adapt to disturbances (such as human behavioral uncertainties or robot actuation noise)  is the key to determining system operational efficiency and safety~\cite{BH-BS:13, CB-EG-OK-MF:20}. One frequently encountered task in engineering settings is object transportation~\cite{SA-KP:22}. To employ a human and a mobile manipulator to perform co-transportation, the key challenges arise from the disturbances~\cite{LS-WH-ZS:21,FC-VM-LL-GO-PK:21, GI-AF-LM:17}, which may not adhere strictly to predefined trajectories, and from the increased control complexity due to the coupling of the robotic arm and its mobile base~\cite{ES-SD-HS:13}. 

To address these challenges, this paper formulates and solves a disturbance-aware Model Predictive Control (MPC) tracking problem enhanced with a pose optimization mechanism. The goal is to derive optimal control strategies by leveraging whole-body control, enabling the robot to estimate the impact of disturbances on costs in terms of tracking errors and energy consumption when simultaneously controlling the mobile base and robotic arm. Building on this, the pose optimization mechanism allows the robot to dynamically adjust its joint angle configurations, more efficiently compensating for trajectory deviations caused by disturbances and reducing control costs.

\subsubsection*{Statement of contribution}
Our contributions in this paper are threefold. Firstly, we formulate a new problem that considers disturbances in an MPC tracking problem that incorporates pose optimization and the robot's whole-body control.
Secondly, we propose a dual-phase optimization strategy, which calculates the estimated control costs in the presence of disturbance within a certain planning horizon, followed by optimization of the robot's pose through selection from a set of joint angle combinations. Lastly, we demonstrate the effectiveness of our method through a combination of theoretical derivation, simulated experiments, and hardware demos using a Fetch robot. We use quantitative comparisons to showcase the advantages of the approach over existing algorithms that do consider disturbances or do not have a pose optimization mechanism.

\section{Literature review} 
\subsubsection*{{Trajectory Tracking}}
There has been considerable existing literature investigating trajectory tracking using a mobile robot or the end effector of a robotic arm. For mobile robots, diverse algorithms such as MPC~\cite{CW-XL-XY-FH-AJ-CY:18}, learning-based nonlinear MPC~\cite{xiao2021learning,CO-AS-TB:14}, sliding mode control~\cite{HT-SR:20}, and adaptive planning can be employed. 
On the other hand, tracking using a robotic arm features algorithms like Gaussian process-based MPC~\cite{AC-EA-MW-LH-MH-MZ:19}, adaptive time delay control~\cite{JB-SC-SH:18}, and model predictive path-following control~\cite{TF-TW-PZ-RF:17}. These works focus on the high degree of freedom and address control problems under complex task configurations and workspace constraints~\cite{FZ-CB-GH:07,AC-NR:20}. 
When combining a robotic arm with a mobile base, only a handful of studies have rigorously addressed their comprehensive whole-body kinematics control~\cite{MG-FF-TS-LS-JB:17,JK-FF-TN-JC-RS-JN:20,MO-MWM-SY-SJ-WM:20}. 
These holistic control mechanisms can significantly expand the operational workspace of the mobile manipulator. Nonetheless, most existing works often do not consider disturbances, particularly the ones introduced by human uncertainties in collaborative tasks.

\subsubsection*{Pose Optimization}
As part of end-effector control with a high degree of freedom, our pose optimization arises from the concept of redundancy resolution~\cite{cavacanti2017redundancy, kim2012redundancy}, where the same end-effector position can be achieved by multiple alternative joint angle combinations.
Redundancy resolution has been studied for robotic arm manipulation using Monte Carlo tree search~\cite{zhang2017active}, stiffness-model of end-effector~\cite{schneider2015automatic}, and sampling-based method~\cite{WJ-ZT-WY-LD:23}. 
Its main benefits include avoiding singularities~\cite{LH-LB:08}, improving reachability in constrained spaces~\cite{HC-ME-TD:21}, and enhancing control accuracy~\cite{AM-BS-LW-LG:14}.  These redundant robot poses are often used to overcome static or dynamic environmental constraints.
The efforts to integrate pose optimization into co-manipulation problems are limited~\cite{MSA-RR:21}, and, similar to the tracking problem, they are not usually designed to better compensate for human uncertainties. 
In contrast, our formulation integrates pose selection with a disturbance-aware MPC formulation, leading to the best pose that minimizes future expected costs.

\subsubsection*{Human-Robot Collaboration}
Adapting robot responses to align with human uncertainties during collaborative tasks is necessary and presents challenges in terms of control efficiency and safety. For this purpose, the integration of reinforcement learning and model-based control has been substantially used, such as using a robotic arm to assist humans in specific target activities through model-based reinforcement learning \cite{LR-JM-PF-AA-FB-LT-ND:20}, wood sawing and surface polishing \cite{LP-NT-DC-AA:17}, engaging in collaborative assembly tasks using Gaussian Process MPC \cite{KH-CH-LP:22}, robust variable admittance control with unknown payload~\cite{mujica2023robust}, sampling-based MPC, such as MPPI, for task and motion planning~\cite{zhang2024multi}, handling task and joint space constraints~\cite{bhardwaj2022storm} etc. Some studies have considered the whole-body kinematics of the mobile manipulator during co-manipulation or transportation tasks with reactive controller~\cite{de2023human}, adaptive collaborative interface\cite{DS-AG-AA:23}, environmental awareness module~\cite{DS-IO-JG-AA:23}; as well as human-humanoid collaboration to reduce human workload~\cite{bussy2012proactive}, carry objects~\cite{agravante2019human, berger2013inferring}, etc. While reinforcement learning shows effectiveness in handling unmodeled human uncertainties, it usually lacks transparency to theoretically ensure performance guarantees. To address this, robust MPC provides a control theoretic approach to address uncertainties in MPC problems, including the consideration of strict safety~\cite{ZM-GS:20} and physical constraints~\cite{SMB-OL-LJH-WS:18}, and systems with varying parameters~\cite{MJ-SCW:12}. 
However, 
they do not consider integrating the MPC formulation with pose optimization to further improve control performance, which is a key difference from the problem considered in this paper.

\section{Preliminaries and Problem Formulation}\label{sec_PF}
We first introduce a trajectory with human uncertainties to which the robot must follow and adapt. Then, we detail the whole-body control of the mobile manipulator, composed of both the mobile base and the robotic arm. Lastly, we mathematically define an MPC tracking problem that is aware of disturbances and incorporates pose optimization.

\textit{Notations: } 
Let $I_r$ denote the $r\times r$ identity matrix; $\text{diag}\{a_1,\cdots,a_r\}$ denote a diagonal matrix. 
For a vector $x$, $|x|_2$ denotes its 2-norm.  For a square matrix $M$, $\tr(M)$ denotes its trace. We use $M\succ 0$ ($M\succeq 0$) to denote the matrix is positive (semi-) definite. We let $\|x\|_{M\succeq0}^2 = x^{\top}Mx$.
\subsection{Trajectory Prediction with Human Uncertainty}
As illustrated in Fig.~\ref{fig_task}, the task is to enable a team comprising a mobile manipulator and a human to collaboratively transport an object. 
We use ${r}=[ r^x, r^y, r^z, r^\alpha, r^\beta, r^\gamma]^{\top}  \in \mathbb{R}^6$ to represent the manipulator's end effector pose to be tracked in order to adapt to human movements, where the entries correspond to three-dimensional positions and orientations (roll, pitch, yaw), respectively. A trajectory is then defined as $r(k)$, $k\in\{0,1,2,\cdots,H\}$ with $H$ being the control time horizon for future time steps.
In general, the trajectory $r(k)$ can be obtained by predicting future human behaviors based on their historic poses. 
Additionally, our formulation considers a disturbance term $\varpi(\tau)$ in the trajectory caused by human uncertainty, which yields
\begin{align} \label{eq_true_traj}
r(k) = \tilde {r}(k) + D\sum_{\tau=0}^k\varpi(\tau).
\end{align}
Here, $\tilde {r}(k)$ is the predicted human behavior, which can be obtained by various different methods in the literature, such as direct curve fitting~\cite{parque2020smooth}, utility theory~\cite{sun2019interpretable}, recurrent neural network~\cite{zhang2020recurrent, kratzer2020prediction} etc. 
If the co-transportation task has a desired nominal trajectory, $\tilde {r}(k)$ may also represent such trajectory that is known to the robot. 
The term $\varpi(\tau)\sim \mathcal{N}(\bm{0}, \Sigma)$ represents the human positional uncertainty that functions as disturbances to $r(k)$.
The distribution of $\varpi$ is assumed to be zero mean and follows a covariance matrix $\Sigma\in\mathbb{R}^{3\times 3}$ in $x,y,z$ directions, where $\Sigma$ encapsulate individual variations of humans. Different values of $\Sigma$ can represent different human preferences, as they tend to introduce disturbances in varying directions and magnitudes.
Given the positional disturbances caused by humans, we assume the desired roll, pitch, and yaw do not change. Thus, $D=\begin{bmatrix}
    I_{3}& \bm{0}_{3\times 3}
\end{bmatrix}^{\top}\in\mathbb{R}^{6\times 3}$. 
In addition, as will be discussed at the end of the next subsection, $\varpi(\tau)$ can also be equivalently used to represent the robot’s actuation disturbances if an appropriate covariance matrix $\Sigma$ can be quantified for the robot.


Since this paper primarily focuses on how robots can adapt to human uncertainties rather than on human modeling itself, in the following, we assume $\tilde {r}(k)$ and $\Sigma$ are known. Note that the use of different methods to obtain $\tilde {r}(k)$ and $\Sigma$ does not impact the correctness of our results.

\subsection{Whole-body Control of a Mobile Manipulator.}
We use a Fetch robot~\cite{Fetch} to present modeling details. However, a similar mechanism is generalizable to a wide class of mobile manipulator platforms.\footnote{Here, for simplicity, we only present the kinematics considering joint velocities. A dynamics model with torque control is given in the supplementary material~\cite{supp}. The following results are directly applicable to the dynamics model.}

\begin{figure}[t]
    \centering
    \includegraphics[height=2.8 cm]{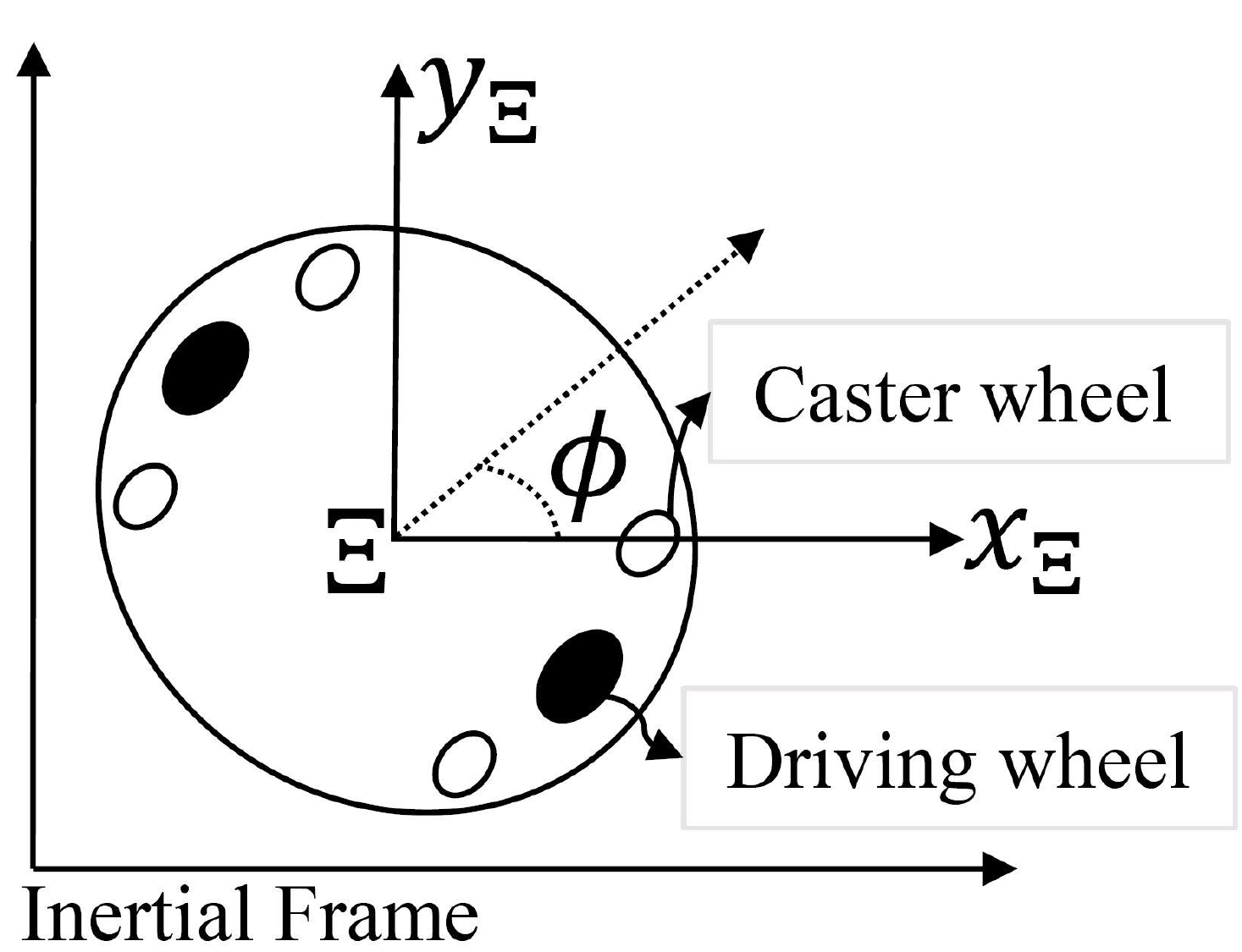}
    \caption{ Fetch robot mobile base in the inertial frame. A shifted frame $\Xi$ based on the current robot position.\vspace{-1.5em}}
    \label{fig_base_wheel}
    \vspace{-0.2cm}
\end{figure}

\noindent \textbf{Mobile Base.}  
As shown in Fig.~\ref{fig_base_wheel}, the base of the Fetch  robot follows a differential drive model:
\begin{align} \label{eq_dynbase}
    s_{\text base}(k+1) =s_{\text base}(k) + \tau \begin{bmatrix}
    \cos (\phi(k)) & 0 \\
    \sin (\phi(k))  & 0 \\
    0 & 1
\end{bmatrix} u_{\text base}(k)
\end{align}
where $s_{\text base}=[x_{\text base},~y_{\text base},~\phi]^{\top}\in\mathbb{R}^3$ represents the $x$, $y$ positions and the heading angle of the robot base in the inertial frame, respectively; $u_{\text base}=[v ~ \eta]^{\top}$ represents the linear and angular velocities in its body frame, $\Xi$. The $\tau$ is the time interval.

\begin{figure}[t]
    \centering
    \includegraphics[width=0.46\textwidth]{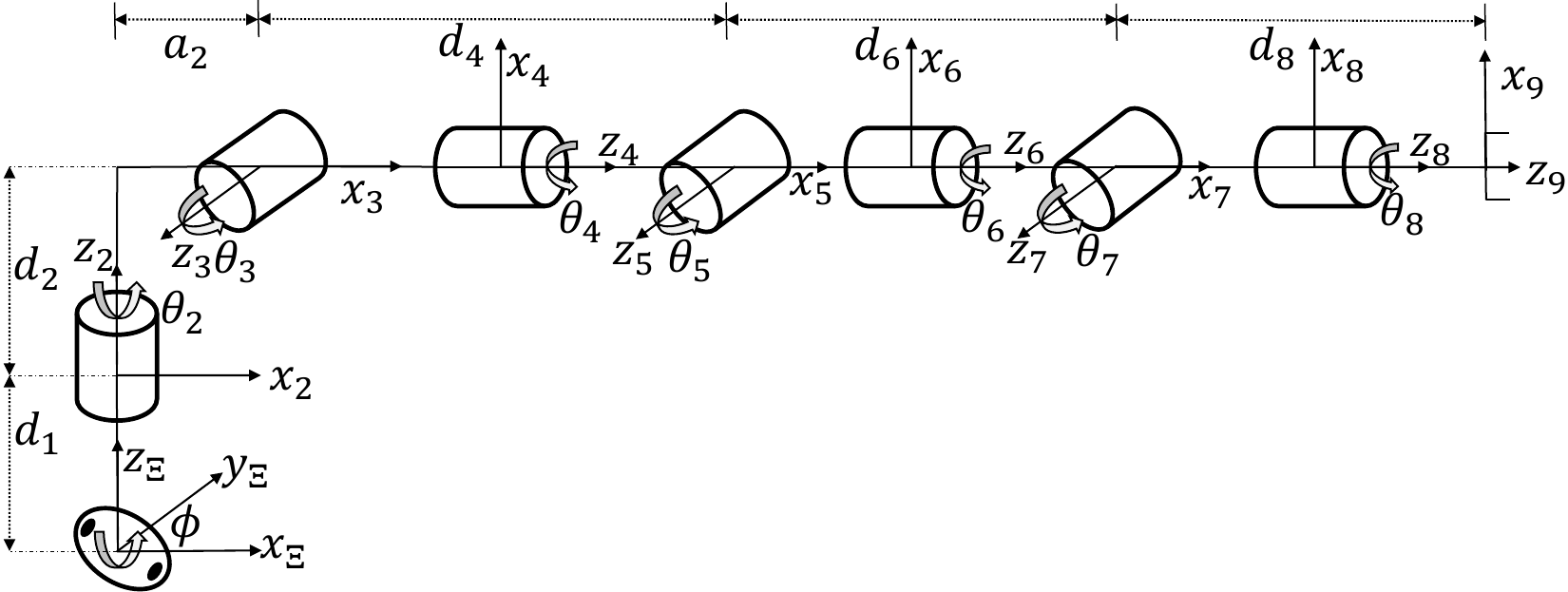}
    \caption{Joint axes configurations of the 8-DOF robotic arm with DH parameters in frame $\Xi$.\vspace{-1.5em}}
    \label{fig_axes_fmr}
    \vspace{-0.2cm}
\end{figure}

\noindent \textbf{Robotic Arm.}
The Fetch robot has a 7-DOF robotic arm built on its mobile base. 
Here, we consider the robot base heading angle $\phi$ as one extra freedom for the robotic arm, which leads to an 8-DOF shown in Fig.~\ref{fig_axes_fmr}. This is for the ease of combining the kinematics of the base and robotic arm, where the end-effector pose in the {$\Xi$} coordinate can be transformed from the inertial frame without rotation. As will be shown later, it also allows us to incorporate angle $\phi$ into \textit{pose optimization}, together with other joint angles.

We represent the end-effector pose of the robotic arm in {$\Xi$} frame by $s_{\text{arm}}=[{p_{\text{arm}}}^{\top}~ {\psi_{\text{arm}}}^{\top}]\in \mathbb{R}^6$, where $p_{\text{arm}} \in \mathbb{R}^3$ denotes the end-effector position in Cartesian coordinates, and $\psi_{\text{arm}}\in \mathbb{R}^3$ denotes the end-effector orientation in Euler angles. 
To study the state transition of $s_{\text{arm}}$, we linearize its forward kinematics equation $f(\cdot)$ using the Jacobian matrix $\mathbf{J}(\cdot)\in\mathbb{R}^{6\times 8 }$ derived based on the DH-parameters~\cite{Fetch} of the robotic arm configured in Fig.~\ref{fig_axes_fmr}: 
\begin{align}\label{eq_dynarm}
    s_{\text{arm}}(k+1) & = {f}(\theta(k)+\tau \omega(k)) \nonumber \\
    & \approx s_{\text{arm}}(k) + \tau \mathbf{J}(\theta(k)) \omega(k) 
\end{align}
where $\theta= [\phi,~ \theta_2,~ \theta_3,\cdots, \theta_8]^{\top}\in\mathbb{R}^8$ includes the mobile base heading angle $\phi$ and the seven robotic arm angles; and 
$\omega = \dot \theta=[\eta, ~\dot\theta_2, ~\dot\theta_3, \cdots , ~\dot\theta_8]^{\top}\in\mathbb{R}^8$ is the corresponding angular velocities. The Jacobian matrix is computed by $\mathbf{J}(\theta(k))=\left.\frac{\partial {f}(\theta)}{\partial \theta}\right|_{\theta=\theta(k)}$. The $\tau$ is the time interval.

\noindent \textbf{Whole-Body Kinematics.}
We combine the kinematics \eqref{eq_dynbase} for the base and \eqref{eq_dynarm} for the robotic arm to obtain the linearized whole-body kinematics of the robot's end-effector pose in the inertial frame as
$s \in \mathbb{R}^{6}$, and
$$s=s_{\text{arm}}+\begin{bmatrix}
    x_\text{base}\\
    y_\text{base}\\
    0_{4\times 1}
\end{bmatrix},$$
where the two states can be directly added because $s_{\text{base}}$ is defined in the inertial frame, $s_{\text{arm}}$ is defined in the $\Xi$ frame, and no rotation is needed for the transition between the two frames.
Consequently, the state update is given by
\begin{align} \label{eq_state_update}
 s(k+1) &= s(k) + B(\theta(k)) u(k)
\end{align}
with
$$B(\theta(k))= \tau \begin{bmatrix}
\renewcommand\arraystretch{1.3}
\begin{array}{c c c c}
    \cos(\phi(k))  \\
     \sin(\phi(k)) & \mathbf{J}(\theta(k))\\
     0_{4 \times 1}     
\end{array}
\end{bmatrix}$$
where $u=[v, ~\omega^{\top}]^{\top}\in\mathbb{R}^{9}$ is the control input combining the linear velocity of the base and all rotations of the robot. $B(\theta(k))\in\mathbb{R}^{6 \times 9}$ is the input matrix depending on the joint angle combinations $\theta(k)$, which combines~\eqref{eq_dynbase} and~\eqref{eq_dynarm}. The second column of~\eqref{eq_dynbase} is integrated into $\mathbf{J}(\theta(k))$. 

Note that if the model \eqref{eq_state_update} is subject to actuation disturbances following a Gaussian distribution, these disturbances can be mathematically merged with $\varpi(\tau)$ without introducing an additional variable, since our goal is to use \eqref{eq_state_update} to track the trajectory \eqref{eq_true_traj}.


\subsection{Disturbance Aware MPC with Pose Optimization} \label{subsec_problem_formulation}

Our research problem focuses on effectively adapting to disturbances, subject to the whole body kinematics of the robot, as per Equation~\eqref{eq_state_update}. 
We solve this by introducing a special MPC formulation, which features a \textbf{pose optimization} mechanism.
Specifically, in addition to the regular control inputs $u(k)$, we allow the robot to change its joint angle combination from $\theta_0$ to a new combination $\bar \theta$ before the execution of control inputs, 
if the new $\bar \theta$ leads to a lower predicted cost over the planning horizon. 
The rationale behind this stems from redundant solutions that given the same end-effector pose realization, an 8-DOF robot can have infinitely many feasible joint angle combinations. If the robot is informed by the future trajectory and the human uncertainty distribution, it can choose a better $\bar \theta$, i.e., a pose with the advantage of inducing less future cost.

The disturbance-aware MPC tracking with pose optimization can be formulated as follows:
\begin{align} \label{eq_problem_formulation}
    & \min_{\bm{u}(0:H-1), \bar\theta \in \Theta} \mathcal{J}(\bm{u}(0:H-1), \bar \theta)\nonumber\\
    \triangleq &~\mathbb{E}_{\varpi} \left[ \sum_{k=0}^{H} \begin{bmatrix}
    s(k) -  r(k) 
    \end{bmatrix}^{\top} Q \begin{bmatrix}
    s(k) -  r(k) 
    \end{bmatrix}\right] \nonumber\\ 
    &~+{\sum_{k=0}^{H-1}} u(k)^{\top} R u(k) + \kappa |\bar \theta - \theta_0|_2^2  \\ 
    &\text{s.t.} \quad s(k+1) = s(k) + {B}(\bar\theta) u(k),~~s(0)=s_0\nonumber\\
    &~\qquad r(k) = \tilde {r}(k) +  D\sum_{\tau=0}^k\varpi(\tau), ~~ \varpi \sim \mathcal{N}(\bm{0}, \Sigma)\nonumber
\end{align}
where $\bm{u}(0:H-1)=\{u(0),\cdots,u(H-1)\}$,
$Q\in\mathbb{R}^{6\times 6}\succeq 0$, $R\in\mathbb{R}^{9 \times 9}\succ 0$ are the weighting matrices for tracking and input costs, respectively; $\kappa\in\mathbb{R}_{+}$ is the cost weight for pose optimization. $s_0$ is the current end-effector pose to initialize each planning phase, and $\theta_0$ is the current joint angle combination. 
For tractability, similar to prior works~\cite{HZ-RRN-GL:14, QS-JZ-AK-ILJ:21}, we consider the linearized system with
a fixed $B(\cdot)$ matrix for the robot's end-effector kinematics for each planning horizon. The impact is small when the planning horizon is short~\cite{QS-JZ-AK-ILJ:21}.

\section{Main Result}\label{sec_main}

To solve problem~\eqref{eq_problem_formulation}, we observe that the optimal control input sequence, $\bm{u}(0:H-1)$, is influenced by the input matrix $B(\bar\theta)$. Thus, it also depends on the robot pose $\bar\theta$, which leads to our pose optimization. Since there are no closed-form solutions for $\bm{u}(0:H-1)$ for general tracking problems in MPC, optimizing both $\bm{u}(0:H-1)$ and $\bar\theta$ at the same time presents a challenge. 
Our approach to address this is a dual-phase method. First, we generate a set of candidate joint angle combinations or $\bar\theta$ values. For each $\bar\theta$, we theoretically compute optimal control inputs, $\bm{u}(0:H-1)$ and estimate the cost-to-go associated with it, considering human uncertainties $\varpi$. 
Then, in the second step, we go through candidate joint angle combinations and choose the one that results in the lowest estimated cost. This combines the best of both tracking cost optimization and pose optimization to find the most efficient $\bar\theta$ and $\bm{u}(0:H-1)$.

We start by presenting the result to solve the optimal control input sequence $\bm{u}(0:H-1)$ and the optimal cost $\mathcal{J}^*$ with a fixed $\bar\theta$. For presentation simplicity, let $\bar{B}={B}(\bar{\theta})$, and define the following error dynamics for~\eqref{eq_state_update} by subtracting $r(k+1)$ from both sides of the equation:
\begin{align}\label{eq_dynerr}
     e(k+1) &= e(k) + \bar B u(k) + r(k) - r(k+1)\\\nonumber
    &= e(k) + \bar B u(k) +  \tilde r(k) -  \tilde r(k+1)- D\varpi(k+1)
\end{align}
with $e(k)=s(k)-r(k)$ being the tracking error.
We hypothesize that the optimal cost-to-go function follows:
\begin{align} \label{eq_cost_function}
    \mathcal{J}^*(e(k), k) = \|e(k)\|^2_{P(k)} + 2e(k)^{\top} p(k) + c(k)
\end{align}
where, $P(k) \in \mathbb{R}^{6 \times 6}, p(k) \in \mathbb{R}^{6}, c(k)\in \mathbb{R}$, are unknown quantities to be determined.
The following result shows that the assumed solution form is valid, and the parameters can be computed from a Discrete Algebraic Ricatti Equation (DARE). The proof of the theorem is given in the Appendix.

\begin{theorem}\label{Thm_1}
Given a $\bar{B}$, assuming the optimal solution $\bm{u}^*$ of~\eqref{eq_problem_formulation} yields an optimal cost $\mathcal{J}^*$ with the form of~\eqref{eq_cost_function}. Then $P(k)$, $p(k)$, and $c(k)$ can be computed by the following disturbance-aware DARE:
\begin{subequations}\label{eq_Para}
\begin{align}\label{eq_P}
     P(k) &=  Q + P(k+1) - P(k+1)\bar B M P(k+1)
     \\[1ex] \label{eq_p}
         p(k) &= p(k+1) +  P(k+1) (\tilde r(k) -\tilde r(k+1))\nonumber \\
         & - P(k+1) \bar B M P(k+1) (\tilde r(k) -\tilde r(k+1))\nonumber \\
         & - P(k+1) \bar B M p(k+1)
         \\[1ex] \label{eq_c}
         c(k)&= c(k+1) +\| \tilde r(k) - \tilde r(k+1)\|^2_{P(k+1)} \nonumber \\
        & +\tr(\Sigma D^{\top}P(k+1)D) -\|P(k+1)(\tilde r(k) \nonumber \\
        & - \tilde r(k+1))+p(k+1))\|^2_{BM} + 2 (\tilde r(k) \nonumber \\
        &- \tilde r(k+1))^{\top} p(k+1)
\end{align}
\end{subequations}
with $M = (R+\bar B^{\top} P(k+1) \bar B)^{-1} \bar B^{\top}$, terminal conditions:
\begin{align*}
    P(k=H)=Q,\quad
    p(k=H)=0,\quad
    c(k=H)=\kappa |\bar \theta - \theta_0|_2^2
\end{align*}
The corresponding control input
\begin{align} \label{eq_control}
    u^*\!(k) \!= \!-M \!(\!P(k\!+\!1)(e(k)\!+\!\tilde r(\!k\!)\!-\!\tilde r(k\!+\!1))\!+\! p(k\!+\!1))
\end{align}
gives the cost in \eqref{eq_cost_function} with parameters in \eqref{eq_Para}.
\hfill $\square$

\end{theorem}

\setlength{\textfloatsep}{2pt}
\begin{algorithm2e} [t]
 \caption{Human-disturbance-aware MPC Tracking with Pose Optimization } 
 \SetAlgoLined \label{Alg_1}
        \textbf{Input} Nominal trajectory $\tilde r_t$; human uncertainty covariance matrix $\Sigma$; current joint angles $\theta_{0}$.\\
        Formulate $r(k=0:H)$ based on equation~\eqref{eq_true_traj}.\\
        Create the joint angles candidate set $\Theta$ by sampling around $\theta_{0}$, also add $\theta_{0}$ to $\Theta$.\\
        \For{each $\bar\theta\in\Theta$}{
                 Compute Jacobian matrix, $\mathbf{J}(\bar\theta)=\left.\frac{\partial {f}(\theta)}{\partial \theta}\right|_{\theta=\bar\theta}$.\\
            Compute matrix $\bar{B}=B(\bar\theta)$ with~\eqref{eq_state_update}. \\
            Solve the MPC by computing solutions for the disturbance-aware DARE in \textit{Theorem 1}. \label{step_solve_DARE}\\
            Compute the optimal cost $\mathcal{J}^*$ associated with the current $\bar\theta$ using~\eqref{eq_cost_function}.}
        Compare the costs $\mathcal{J}^*$ for all candidate $\bar\theta$ and find the optimal $\bar\theta^*$ as pose optimization. Compute the control for pose optimization as $\bar{\bm{u}}=\frac{\bar\theta^*-\theta_0}{\tau}$. 
 \\   
        Reuse step~\ref{step_solve_DARE} for the selected $\bar\theta^*$ and equation~\eqref{eq_control} to generate control input sequence $\bm{u}^*(0:H-1)$.\\
        Add $\bar{\bm{u}}$ and $\bm{u}^*(0)$ then apply them to the robot.
\end{algorithm2e}

\subsubsection*{Proof of Theorem 1}
Integrating the error update~\eqref{eq_dynerr} into the optimal cost~\eqref{eq_cost_function}, one has
\begin{align}\label{eq_J1}
    &\mathcal{J}^{\star}(e(k),k)=\min_{{u}(k)} \mathcal{J}(e(k),k)) \nonumber\\
    =~ &{\min_{{u}(k)}} [ \mathbb{E}_{\varpi}[\|e(k)\|_Q^2  + \|u(k)\|_R^2 +\mathcal{J}^{\star}(e(k+1)),k+1)]] \nonumber \\
    =~ &{\min_{{u}(k)}} [ \mathbb{E}_{\varpi}[\|e(k)\|_Q^2  + \|u(k))\|_R^2  \nonumber \\
    &+ \|e(k)+ \bar B u(k)+ \tilde r(k)- \tilde r(k+1)\|^2_{P(k+1)} \nonumber \\
    & +  \tr(\Sigma D^{\top}P(k+1)D) + 2 (e(k)+ \bar B u(k) \nonumber \\
    & + \tilde r(k) - \tilde r(k+1) )^{\top} p(k+1)+ c(k+1)]]  
\end{align}
where the elimination of terms follows from $\mathbb{E}(\varpi(k+1))=0$.
Since the control input minimizes cost at each time step,
the optimality condition $\frac{\partial \mathcal{J}^{*}(e(k),k)}{\partial u(k)}=0$ yields
\begin{align} \label{eq_optu}
    &Ru(k)+ \bar B^{\top} P(k+1)  (e(k)+ \bar B u(k)+  \tilde  r(k) - \tilde  r(k+1) ) \nonumber\\
    &\qquad \qquad \qquad +{\bar B}^{\top}p(k+1)=0  
\end{align}
Thus, for each time-step $k$, the control inputs follows:
\begin{align}\nonumber
    u^*(k) \!= \!-M (\!P(k\!+\!1)(e(k)\!+\!\tilde r(k)\!-\!\tilde 
 r(k\!+\!1))\!+\! p(k\!+\!1))
\end{align} 
with $M$ defined in Theorem~\eqref{Thm_1}.
Bring this back to~\eqref{eq_J1} and reusing condition~\eqref{eq_optu} by left multiplying $u(k)^{\top}$ yields
\begin{align} \label{eq_cost_DARE}
    \mathcal{J}^*(e(k),k) &=  \|e(k)\|_{[Q+P(k+1)-P(k+1)\bar B M P(k+1)]}^2 \nonumber \\
    & + 2e(k)^{\top}\big[p(k+1)- P(k+1) \bar B (M p(k+1)\nonumber \\
    & ~~-  P(k+1) \bar B M P(k+1)(\tilde r(k) - \tilde r(k+1)    \nonumber \\
    & ~~+ P(k+1) (\tilde r(k) -\tilde r(k+1))\big] \nonumber \\
    & + c(k+1) +\|\tilde r(k) - \tilde r(k+1)\|_{P(k+1)} \nonumber \\
        & ~~+\tr(\Sigma D^{\top}P(k+1)D) -\|P(k+1)(\tilde r(k) \nonumber \\
        & ~~- \tilde r(k+1))+p(k+1))\|^2_{BM} + 2 (\tilde r(k) \nonumber \\
        &~~- \tilde r(k+1))^{\top} p(k+1)
\end{align}
Comparing~\eqref{eq_cost_DARE} with~\eqref{eq_cost_function}, we have the DARE for $P(k)$, $p(k)$, and $c(k)$. The terminal conditions are obtained by considering $\mathcal{J}^*(e(H), k=H)$ for~\eqref{eq_problem_formulation}, where $P(H)=Q$ and $c(H)$ is the constant pose optimization cost. \hfill\qed

\smallskip
From Theorem~\ref{Thm_1} and equation~\eqref{eq_control}, it can be observed that the optimal control input $u^*(k)$ does not depend on $c(k)$. However, $c(k)$ contributes to the computation of optimal cost $\mathcal{J}^*$, which impacts the pose selection among the candidates.
Specifically, our two-step solver is summarized in Algorithm~\ref{Alg_1}. For each MPC horizon, we first create the joint angles candidate set $\Theta$ by randomly changing multiple joint angles of $\theta_0$ with a small radian value. 
For every $\bar \theta\in\Theta$, we compute the control matrix $\bar B(\bar \theta)$. Then, we use Theorem~\ref{Thm_1} to compute the estimated optimal cost $\mathcal{J}^*$ for the current $\bar \theta$. This process is parallelizable to improve computational efficiency. By exploiting all $\bar \theta\in\Theta$ and the associated $\mathcal{J}^*$, we select the best $\bar \theta$ and use~\eqref{eq_control} to obtain the associated control inputs. 
For implementation, to precisely follow our definition of control cost, we should first apply pose optimization and then apply arm control $\bm{u}^*$. However, this separation can be challenging for the hardware runtime. To address this, we fuse the pose optimization (if exists) and the arm control inputs into a single step. Based on the triangle law of vector addition, the combined implementation will always result in no greater cost than the separated implementations. Furthermore, the cost difference is minimal when the angle change for pose optimization is sufficiently small.

\begin{figure*}[t]
    \centering
    \includegraphics[width=0.9\textwidth]{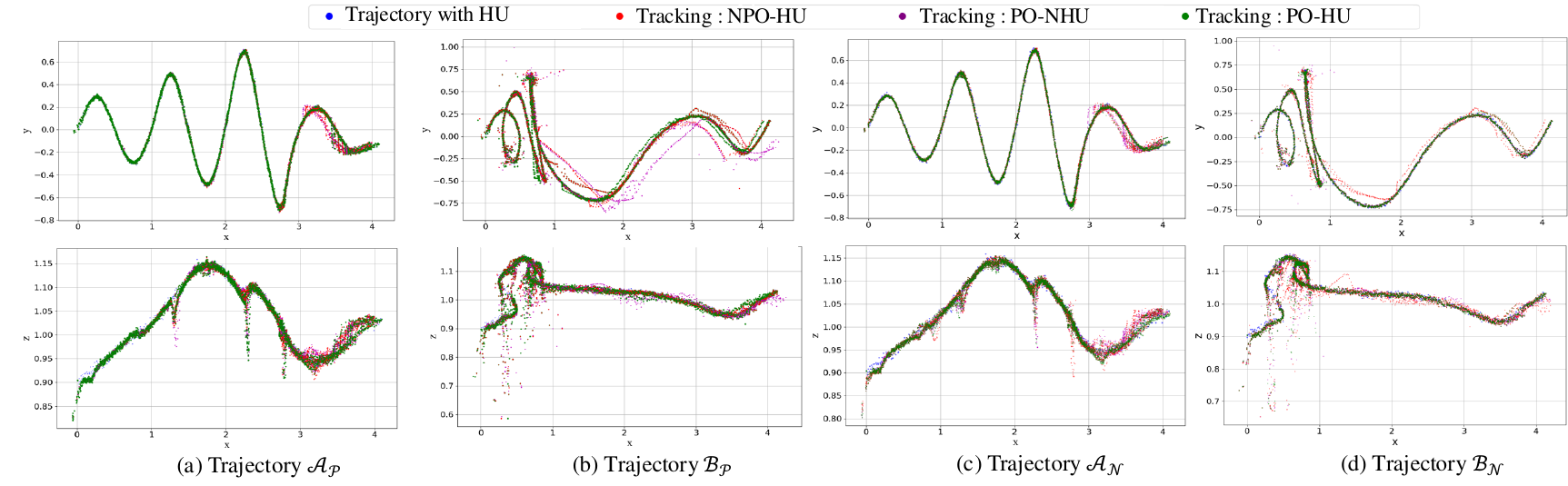}
    \caption{All trajectories with human uncertainties, $Q = 1000\cdot I_6$, $q = 0.4$. The results of trials are represented by dot clouds.}
    \label{fig_all_traj}
    \vspace{-0.7cm}
\end{figure*}
\begin{remark}
In general, the highly non-linear relation between $B(\bar\theta)$ and $\bar\theta$ makes it computationally infeasible to systematically find the optimal pose $\bar\theta$ for estimated control cost $\mathcal{J}^*$. Instead, in Algorithm~\ref{Alg_1}, we employ a sample-based approach to select a candidate set for pose optimization. This allows us to numerically search for a pose $\bar\theta$ that is better than $\theta_0$ in terms of future control costs. A similar technique has been used in~\cite{WJ-ZT-WY-LD:23}. Increasing the cardinality of $\Theta$ can potentially lead to a better $\bar\theta^*$ using more computation, and the evaluation of $\bar\theta$ can be performed in parallel. 
Furthermore, as we will demonstrate in experiments, 
optimizing poses periodically over fixed intervals (not every step) can also improve tracking performance.
Lastly, to make the pose optimizing more efficient, one may leverage deep learning methods to determine when a pose optimization is needed~\cite{LJ-RF-YM:22}, and how a candidate set should be chosen~\cite{LT-SK-JY-LH:11}.

\end{remark}

\section{Experiments}

In this section, we evaluate our proposed algorithm through simulation experiments in Gazebo and a demonstration using a fetch robot~\cite{Fetch}. 
In the following, we assume human uncertainties are the source of disturbance, which is mathematically equivalent if the robot has actuation disturbances. 
To simulate human uncertainties, we assume  $\varpi(k) \sim \mathcal{N}(\bm{0}, \Sigma)$, where $\Sigma = q \cdot \text{diag}(0.015, 0.025, 0.015)$ (meters) and $q\in  \{0.4,~0.7\}$.
This reflects the tendency for disturbances to be more pronounced along the $y$-axis compared to the $x$ and $z$-axes, with $q$ being the strength of these disturbances.

We test our algorithm by considering two trajectories under two cases. For one case, we assume human future poses are unknown and use LSTM~\cite{nie2017monocular}
for pose prediction to obtain $\tilde{r}$ in \eqref{eq_true_traj}. We use two layers of LSTM, each containing 50 units and a dense layer (output layer). The LSTM layers are fed with the sequences of historic poses, and the output of the layers is the predicted poses.\footnote{Details of LSTM implementation are given in the supplementary material~\cite{supp}.} 
The experiments are denoted as $\mathcal{A_P}$ and $\mathcal{B_P}$. We also test the algorithm assuming human future poses are known as a nominal trajectory $\tilde{r}$, although the trajectory is still subject to human uncertainties. The experiments are denoted as $\mathcal{A_N}$ and $\mathcal{B_N}$.
Each trajectory has 500 discrete points, with a time interval of $\tau = 0.1$ seconds. 
The parameters used for solving the problem~\eqref{eq_problem_formulation}, are selected to be $H = 3$ (MPC horizon) for $\mathcal{A_P}$ and $\mathcal{B_P}$; $H = 8$ for $\mathcal{A_N}$ and $\mathcal{B_N}$. We let $R = I_9$, and $\kappa = 1$. We test two different settings for $Q$, with $Q = 1000 \cdot I_6$ or $Q = 500 \cdot I_6$ to reweigh the tracking error on the overall cost.

Our experiments follow Algorithm~\ref{Alg_1}, and we repeat the MPC planning every time step. In each planning horizon, we chose twelve candidate poses in $\Theta$. While the proposed algorithm seeks to minimize an expected cost, we define the true system cost over the entire trajectory as:
\begin{align*} 
    \mathcal{C}_\text{total} =\sum_{t=1}^{T} e(t)^{\top}Qe(t) + u(t)^{\top}Ru(t) + \kappa|\bar \theta(t) - \theta(t)|_2^2,
\end{align*}
which takes into account the costs for the robot's end effector tracking error, control input, and pose optimization, $T$ is the total number of time steps.
We compare our proposed approach (PO-HU: considering pose optimization and human uncertainty) with two baselines: one approach with No Pose Optimization but considering Human Uncertainty (NPO-HU) and another one with Pose Optimization but Not considering Human Uncertainty (PO-NHU). Note that we do not evaluate the no pose optimization and no human uncertainty case, as it is equivalent to NPO-HU. This equivalence arises because human uncertainty with zero mean does not affect the control inputs (but only the expected control cost), as shown in \eqref{eq_control}.

The dot clouds in Fig. \ref{fig_all_traj} compare tracking performance across trajectories on the x-y and x-z planes, with the weighting matrix $Q = 1000 \cdot I_6$ and $q = 0.4$. 
A numerical comparison is further provided in Table~\ref{table_performance}, which shows the average total cost $\mathcal{C}_{\text{total}}$ of different algorithms over 100 trials. 
Additionally, we also introduce a periodic pose optimization (pPO-HU) that performs pose optimization every 5 time steps. This helps to reduce the computational burden when applied to low-cost devices.
For hardware validation, all the trajectories computed from Gazebo visualized in Fig. \ref{fig_all_traj} have been executed and successfully reproduced on our hardware platform to justify their feasibility.


\begin{table}[h]
\caption{Comparison of $\mathcal{C}_{\text{total}}$ across Different Algorithms}\vspace{-1em}
\begin{center}
\begin{tabular}{|m{0.4cm}|m{0.68cm}|m{0.22cm}|m{0.92cm}| m{1.07cm}| m{1.11cm}| m{1.11cm}|}
    \hline
        \textbf{Traj} & $\mathbf{Q}$ & $\mathbf{q}$ & \textbf{PO-HU} & \textbf{pPO-HU} & \textbf{NPO-HU} & \textbf{PO-NHU} \\
    \hline
        \multirow{4}{*}{$\mathcal{A_P}$} & \multirow{2}{*}{$1000I_6$} & 0.4 & \textbf{744.04}  & 841.06  & 2364.77$^\dagger$ & 1339.781 \\
        \cline{3-7}
        & & 0.7 & \textbf{1273.42} & 1303.88 & 2394.74 & 2427.38 \\
        \cline{2-7}
        & \multirow{2}{*}{$500I_6$} & 0.4& \textbf{846.64} & 847.77  & 901.58 & 898.16 \\
        \cline{3-7}
        &  & 0.7&  \textbf{999.66} & 1003.27 & 1077.64 & 1056.52 \\   
    \hline
        \multirow{4}{*}{$\mathcal{B_P}$}& \multirow{2}{*}{$1000I_6$} & 0.4 & \textbf{2436.97} & 2671.56 & 5546.44$^\dagger$ & 3172.04\\
        \cline{3-7}
        & & 0.7 & \textbf{3831.34} & 4083.78  & 9731.61$^\dagger$ & 9234.85$^\ddagger$ \\
        \cline{2-7}
        & \multirow{2}{*}{$500I_6$} & 0.4& \textbf{2030.57} & 2138.46  &8752.65$^\dagger$  &2409.09 \\
        \cline{3-7}
        &  & 0.7&  \textbf{2324.85} & 2499.22 &9589.51$^\dagger$  & 7509.28$^\ddagger$ \\
    \hline
        \multirow{4}{*}{$\mathcal{A_N}$} &\multirow{2}{*}{$1000I_6$} & 0.4 & \textbf{686.94} &769.55 &   1932.83$^\dagger$ & 1102.95 \\
        \cline{3-7}
        & & 0.7 & \textbf{1151.33} & 1292.57  & 2342.11$^\dagger$ & 2384.16$^\ddagger$\\
        \cline{2-7}
        &  \multirow{2}{*}{$500I_6$} & 0.4& \textbf{829.59} & 835.64  & 851.60 & 852.53\\
        \cline{3-7}
        &  & 0.7& \textbf{978.52} & 986.46  & 1013.69 & 1017.27\\
    \hline
        \multirow{4}{*}{$\mathcal{B_N}$}& \multirow{2}{*}{$1000I_6$} & 0.4 & \textbf{2306.64}   &  2540.73 & 3759.09$^\dagger$ & 2968.25\\
        \cline{3-7}
        & & 0.7 & \textbf{3782.70} &3943.75&  8694.57$^\dagger$  & 7220.29$^\ddagger$\\
        \cline{2-7}
        & \multirow{2}{*}{$500I_6$} & 0.4& \textbf{2013.31} &   2041.27  & 2261.32 & 2209.01\\
        \cline{3-7}
        &  & 0.7&  \textbf{2293.11}  & 2481.39   & 5612.76$^\dagger$  & 4301.46$^\ddagger$\\
    \hline
\end{tabular}
\label{table_performance}
\end{center}
\vspace{-1em}
\end{table}

The effectiveness of the proposed algorithm is justified by Fig.~\ref{fig_all_traj}. In all cases, the result of PO-HU closely follows the trajectories, whereas the other two baseline algorithms deviate a lot. This is further reinforced by the numerical results in Table~\ref{table_performance}.
Specifically, in the table, we use $\dagger$ in NPO-HU, and $\ddagger$ in PO-NHU to highlight the entries where the differences are significant. The fewer highlighted entries in PO-NHU imply that pose optimization has a greater impact on the total cost than the characterization of human uncertainty. 
Additionally, when performing pose optimization every five time steps (pPO-HU), it performs better than the other two, although having a small gap compared to performing pose optimization at every time step (PO-HU). 
Finally, from both figures and tables, we observe that in general, the robot performs better in $\mathcal{A_N}$, $\mathcal{B_N}$. This is because robots have access to nominal trajectories of human future poses, which is more informative than human prediction cases $\mathcal{A_P}$, $\mathcal{B_P}$. This is especially true for trajectory $\mathcal{B}$ because it is more complex and has sharp turns, where the predicted human future poses can be inaccurate.

\begin{figure}[h]
    \centering
    \includegraphics[width=0.44\textwidth]{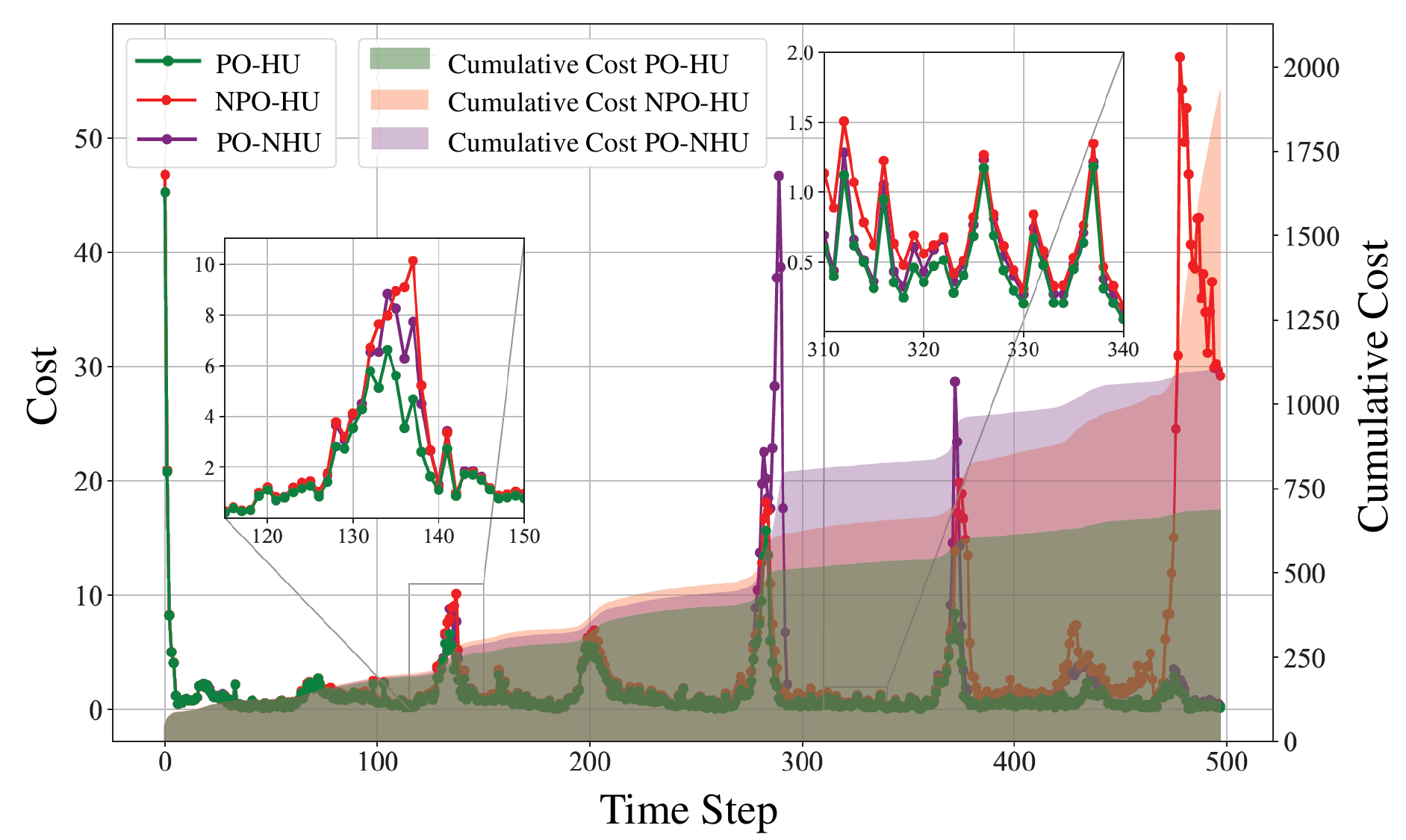}
    \caption{Accumulated cost comparison for trajectory $\mathcal{A_N}$, with $Q = 1000 \cdot I_6$, $R = I_9$, $q = 0.4$. \vspace{-0.5em}}
    \label{fig_merged_cost_plot}
\end{figure}

Fig.~\ref{fig_merged_cost_plot} visualizes the accumulated cost over time and the cost at each time step for trajectory $\mathcal{A_N}$. 
A comparison of cost and peaks reveals the effectiveness of our algorithm in navigating complex trajectory segments, particularly during sharp turns. Furthermore, it is observed that towards the end of the trajectory, the costs associated with NPO-HU explode because the robot admits a bad pose and can hardly reach and track the remaining trajectory. In contrast, the costs for the methods with pose optimization remain relatively stable.
Similar trends are seen for other cases.

\begin{figure} [t]
    \centering
    \includegraphics[width=0.44\textwidth]{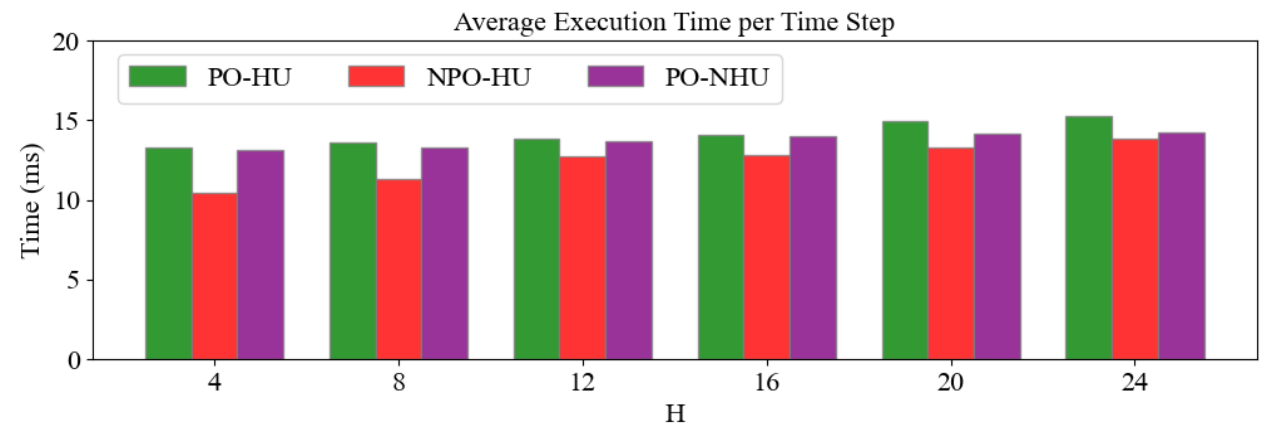}
    \caption{Execution time comparison with different planning horizon, $|\Theta|=12.$ Test computer uses AMD 5975XW.} 
    \label{fig_execution_time}
\end{figure}

Fig.~\ref{fig_execution_time} shows the average execution time of the proposed algorithm for different planning horizons, averaging over 500 trials. Steps (4-9) of the proposed algorithm can be fully parallelized, thus the computation time for pose optimization only increases slightly, as long as the size of the set $|\Theta|=12$ is smaller than the number of computing threads. 
Furthermore, the execution time for each algorithm increases only mildly across different values of $H$. The main computation time is spent on the one-time computation of the Jacobian matrix. The minor increase is due to more iterations in solving the DARE (cf. equation \eqref{eq_Para}). This concludes that our proposed PO-HU method does not introduce significant extra execution time compared with other baseline algorithms.

\section{Conclusions and Future Works}
We studied the control of a mobile manipulator to perform human-robot co-transportation tasks. By modeling human uncertainties and the whole-body kinematics of the robot, we formulated a new human-disturbance-aware MPC tracking problem with pose optimization. 
We proposed an algorithm with a two-step iterative design, equipped with an inner loop that computes an disturbance-aware DARE to estimate the control cost, and an outer loop that selects the best pose with the minimum cost from a candidate set. The correctness and effectiveness of the proposed approach have been validated both theoretically and experimentally. 
Future work will include the generalization of the algorithm to multi-human multi-robot collaborative tasks.

\bibliography{bibliography}
\bibliographystyle{ieeetr}

\end{document}